\documentclass[journal]{IEEEtran}
\usepackage{indentfirst}
\usepackage{amsmath}
\usepackage{graphicx}%调用graphicx包
\usepackage{booktabs}
\usepackage{algorithm}
\usepackage{algpseudocode}
\usepackage{amsmath}
\usepackage{subfigure}
\usepackage{cite}
\usepackage{amssymb}
\ifCLASSINFOpdf
\else
\fi
\begin{document}
\title{Target Search and Navigation in Heterogeneous\\Robot Systems with Deep Reinforcement Learning}

\author{Yun~Chen, Jiaping~Xiao
\thanks{Y. Chen is with the Department of Mechanical Engineering, University of Hong Kong. J. Xiao is with the School of Mechanical and Aerospace Engineering, Nanyang Technological University, Singapore 639798, Singapore (Corresponding author: Jiaping Xiao, e-mail: jiaping001@e.ntu.edu.sg).}}

\maketitle

\begin{abstract}
Collaborative heterogeneous robot systems can greatly improve the efficiency of target search and navigation tasks. In this paper, we design a heterogeneous robot system consisting of a UAV and a UGV for search and rescue missions in unknown environments. The system is able to search for targets and navigate to them in a maze-like mine environment with the policies learned through deep reinforcement learning algorithms. During the training process, if two robots are trained simultaneously, the rewards related to their collaboration may not be properly obtained. Hence, we introduce a multi-stage reinforcement learning framework and a curiosity module to encourage agents to explore unvisited environments. Experiments in simulation environments show that our framework can train the heterogeneous robot system to achieve the search and navigation with unknown target locations while existing baselines may not, and accelerate the training speed.
\end{abstract}


% \fi
\IEEEpeerreviewmaketitle

\section{Introduction}
\IEEEPARstart{W}{ITH} the development of automation and artificial intelligence, research on mobile robots has made significant breakthroughs and has been applied in various fields. Currently, mobile robots are widely used in search and rescue (SAR) scenarios, as they can help explore complex and unknown environments and improve rescue efficiency while reducing the workload of rescue personnel\cite{liu2013robotic}.

In the environment of underground mines (in the event of a mine accident), most environmental information is unknown, where the vision of mobile robots is obstructed since there are many obstacles, and the underground environment signal is weak, making it infeasible for the human remote control to complete the search and rescue tasks. Therefore, in this environment, robots need to have the ability to autonomously complete tasks. However, in such unknown and complex environments, unmanned ground vehicles (UGV) have significant limitations in localization and poor perception of complex terrain, which can only perform local path planning. These shortcomings can make it difficult to quickly search for victims and carry out rescue operations\cite{niroui2019deep}. On the other hand, unmanned aerial vehicles (UAV) face limitations in endurance and the inability to carry a large amount of equipment. To overcome these challenges, using an aerial-ground robot system to enhance their perception and operation is an effective method. The aerial-ground robot system consists of a UGV and a UAV. Through information exchange and collaborative behavior, it can greatly improve the navigation and obstacle avoidance capabilities of the UGV-only system in complex and unknown environments\cite{liu2022review}.

In general, robot navigation problems involve determining a collision-free path from oneself to the target location while minimizing the cost of the navigation path. Existing optimized navigation methods are commonly classified into global navigation methods and local navigation methods. Common global navigation methods include A*, Rapidly-Exploring Random Tree (RRT), etc., while local navigation methods include Artificial Potential Field (APF) method, dynamic window approach (DWA), etc. Other heuristic methods include Neural Network (NN), fuzzy logic, Genetic Algorithm (GA), etc \cite{gul2019comprehensive}. Global navigation methods require prior knowledge of the whole environment while local navigation methods always require longer computation time.
\begin{figure}[!tbp]
  \centering
  \includegraphics[width=3.4in]{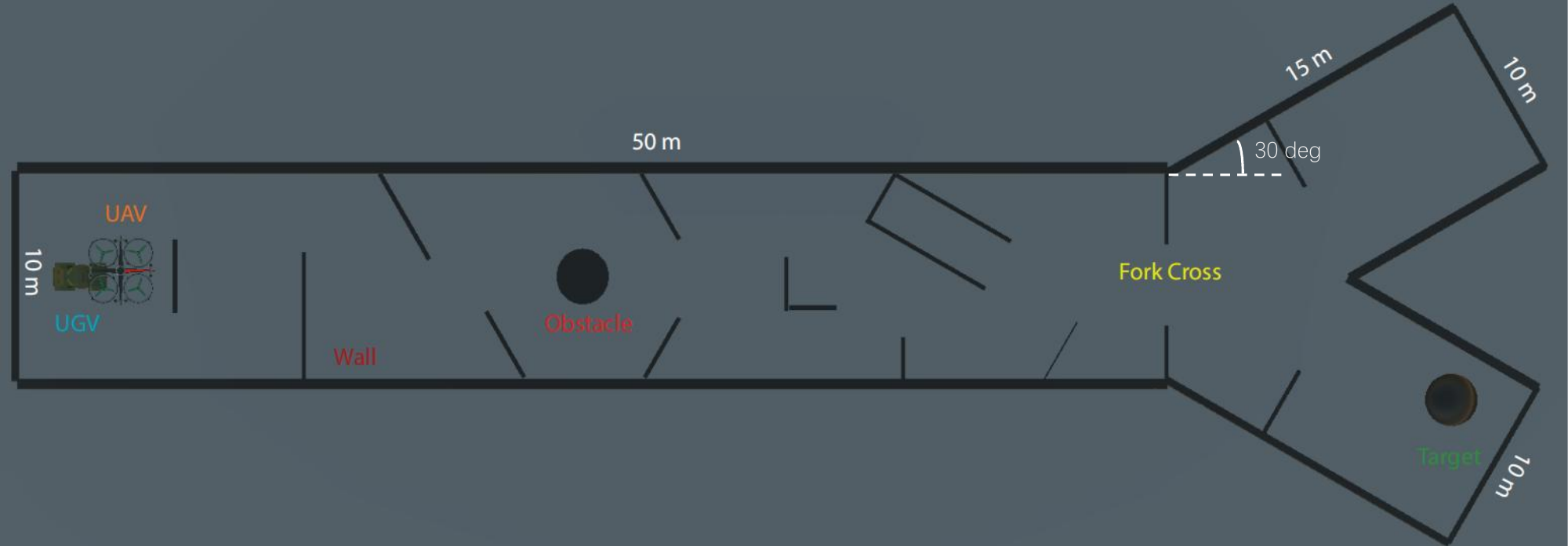}
  \caption{Top view of the designed simulation environment for search and rescue in underground mine scenario. The black lines denote the wall and the sphere-represented victim randomly appears in one of the two branches during the environment generation.}
  \label{fig1}
\end{figure}

 Currently, machine learning algorithms have been widely applied in computer vision\cite{kostrikov2020drlimage}, navigation\cite{liu2021lifelong}, and obstacle avoidance\cite{wu2020towards}, as well as multi-agent systems research\cite{xiao2022collaborative}, and have achieved promising results. Among them, reinforcement learning (RL) enables intelligent agents to learn and interact with the environment and update their policies more effectively based on rewards compared to traditional algorithms. RL has been used in robot search and rescue and navigation, but it faces the curse of dimensionality problem\cite{nguyen2020deep} as the complexity of the environment increases, which limits its application scenarios. The emergence of deep reinforcement learning (DRL) algorithms has solved these problems to some extent with the help of neural networks. State features can be extracted and learned through neural networks, thus dimensions can be reduced to some extent. A commonly used DRL algorithm is Proximal Policy Optimization (PPO)\cite{schulman2017proximal}, which is a policy gradient algorithm that updates policies through gradient ascent and restricts policy variation to reduce variance and convergence time while maintaining efficiency and stability, and achieves good learning results. However, during the training process of DRL, due to the issue of sparse rewards and high-dimensional state space, the policy may have difficulty in converging to the global optimal solution.

To address these challenges, this paper proposes a multi-stage reinforcement learning approach for the collaborative exploration and navigation of heterogeneous robot systems in underground mine environments. The main contributions of this paper are summarized as follows:

1) We propose a learning-based method for collaborative search and navigation of heterogeneous robot systems, which achieve a high success rate in simulated environments. This method uses a multi-stage reinforcement learning approach and trains the policies of UAV and UGV in two stages. In addition, an intrinsic curiosity module (ICM) is introduced to address the issue of sparse rewards, enabling the agents to explore the environment more efficiently during training.

2) The proposed navigation method does not require any mapping, and UGV only needs to follow UAV while avoiding obstacles to reach the target location. Moreover, the UGV and UAV do not obtain any target location-related information during the entire process.

3) We evaluate the performance and generalization capability of our trained policy in increasingly complex environments. The results demonstrate superior performance compared to existing baseline methods.

The remaining part of this paper is structured as follows. In Section II, we summarize the related works. In Section III, we describe the problem of collaborative search and rescue navigation of heterogeneous robot systems. A multi-stage reinforcement learning framework is proposed in Section IV. Section V conducts simulation experimental verification of our method and discusses our results. Section VI concludes our work and provides prospects for future work.

 
\section{RELATED WORKS}
\subsection{Heterogeneous Robotic Systems}
Heterogeneous robot systems have become a hot research topic in the field of robotics. Such systems are typically composed of multiple types of robots, including ground robots, aerial robots, and underwater robots, each of which can perform different tasks or cooperate with each other to improve the overall performance of the system. Many studies have explored the applications of heterogeneous robot systems, such as using drones as communication nodes or messengers for unmanned vehicles\cite{ding2019messenger} or using unmanned ground vehicles as mobile base stations for drone swarms. Additionally, a more collaborative approach would use drones as sensors or decision makers and other autonomous vehicles as actuators\cite{zhang2018uavsensor}.

Furthermore, many researchers have successfully achieved autonomous search and navigation with air-ground robot systems, including cooperative navigation based on map building, as well as cooperative navigation based on state estimation, etc. In \cite{shen2017collaborative}, a method for cooperative search and navigation between aerial and ground robots was proposed. The UAV moved in a lawn-mower pattern to survey an area, recorded the location of the target upon discovery, and then returned to the starting point. Subsequently, the UGV equipped with a LiDAR used EKF-based online path planning to approach the target based on point cloud information, while the UAV ran the same EKF algorithm to follow the UGV until the target point was reached. The results of real-world experiments showed that the system was completely autonomous and had good robustness in different environments. In \cite{mueggler2014aerial}, the UAV flew in a lawn-mower pattern and captured images at fixed locations to build a grid map, based on the path planning performed. The UGV moved along the specified path while adjusting the movable obstacles on the path, ultimately reaching the target point. However, in these works, the target location and map information are required before the UGV can operate, which can affect the speed of rescue. In our work, the UAV and UGV are expected to operate simultaneously to accomplish the required task without the prior target information.

\subsection{DRL-Based Navigation}
The application of DRL in the field of mobile robotics is a valuable research direction. DRL has been used to solve complex problems such as local obstacle avoidance and path planning, exhibiting stronger robustness and scalability compared to traditional algorithms. Depending on the sensors carried by the robots, navigation methods can be classified into vision-based and laser-based 
methods.

A target-driven visual navigation model was proposed in \cite{zhu2017target}, which trained an agent using the actor-critic method in DRL and designed a simulation environment with high-quality 3D scenes and a physics engine Unity3D. This approach improved the generalization of the policy to both the target and the environment, while also enabling a better transfer capability of the policy to the real world.
In\cite{tai2017virtualtoreal}, an asynchronous DRL algorithm is used to train a continuously controlled unmanned vehicle. A mapless navigation strategy requiring only ten sparse laser readings and a target coordinate as input is implemented. This strategy outputs velocity commands based on observations and performs well in both simulation and real-world environments. A method based on the Double Deep Q-Network (DDQN) was presented in\cite{marchesini2020discrete}, and it demonstrated the potential of discrete action space algorithms to replace continuous ones in mapless navigation problems. The authors confirm that the model trained using the ml-agents toolkit provided by Unity can be directly deployed on a real robot without further training.

However, in previous studies, such as \cite{tai2017virtualtoreal}, the observation of the unmanned vehicle contains information related to the target position and cannot be directly applied to search and rescue scenarios. In contrast, in this paper, the unmanned vehicle navigates solely based on following the trajectory of a drone, without any direct observation of the target. This approach addresses various drawbacks of unmanned vehicle search and rescue in complex environments by leveraging the high mobility and visual advantage of the drone to assist the unmanned vehicle in completing the search and navigation task faster. To achieve this goal, a multi-stage reinforcement learning framework is proposed and applied to train a heterogeneous robot system. This approach successfully achieves collaborative search and navigation with high success rates.
\section{UAV-UGV COLLABORATION PROBLEM}
\subsection{Problem Statement}
The designed environment in which the UGV and UAV perform search and navigation tasks is a mine tunnel, as shown in Fig. \ref{fig1}. The tunnel is equipped with dense obstacles and ends with a crossroad. The victim randomly appears in one of the two branches during the environment generation, represented by a spherical object from a top-down view of the environment. It should be noted that the victim is always obstructed by walls in the scene, so the UGV cannot determine the victim's location through laser information and can only navigate by following the UAV. The UAV needs to search for the victim without touching the walls and then approach the victim, while the UGV needs to avoid road obstacles while following the UAV until reaching the target point. The policy of the agents in this task can be described as follows,
\begin{equation}
    \pi_{\text{ugv}}(a_{\text{ugv}}|s_{\text{ugv}}) = P(a_{\text{ugv}}|L_{\text{ugv}}, dir, O_{\text{ugv}}, a_{\text{ugv\_prev}})
\end{equation}

\begin{equation}
    \pi_{\text{uav}}(a_{\text{uav}}|s_{\text{uav}}) = P(a_{\text{uav}}|L_{\text{uav}}, O_{\text{uav}}, a_{\text{uav\_prev}})
\end{equation}
where the $s_i$ and $a_i$ for $i\in\{\text{ugv}, \text{uav}\}$ are the states and actions of agents respectively. $L_i$ is the laser observations while $O_i$ is the ego-state observations. $dir$ is the direction measurements w.r.t. the UAV. Our goal is to train the agents through a reinforcement learning algorithm and find out optimal policy networks $\pi_{\text{ugv}}$ and $\pi_{\text{uav}}$ as shown above.

\subsection{Proximal Policy Optimization 2 (PPO2) Algorithm}
In reinforcement learning, problems can often be formulated as Markov Decision Processes (MDPs). However, in practical situations, the environmental state $S_t$ is not fully observable. Therefore, it is more accurate to describe the problem using a Partially Observable Markov Decision Process (POMDP). A POMDP can be represented as a septuple $(S_t, A_t, P, R_t, O_t, \gamma)$, where the agent receives an observation $O_t$ that contains partial information about the environmental state (e.g., data from a LiDAR sensor), takes an action $A_t$, and obtains a reward $R_t$. Specifically, we can represent the sequence of $(O_t, A_t, R_t)$ by the trajectory $\tau_t$, and the action $A_t$ is generated according to the policy $\pi$ and the observation $O_t$, represented as $A_t \sim \pi(\cdot | O_t)$. The next state $S_{t+1}$ is generated based on the state transition function $T$ (the transition probability $P$ in stochastic policy), the current state $S_t$, and the action $A_t$, represented as $S_{t+1} \sim T(S_t, A_t)$. The optimal policy can be represented by the following formula
\begin{equation}
    \pi_\text{opt} = \text{argmax}  E \left[\sum_t (\gamma^{(t-1)} R_t) | S_0 \right]
\end{equation}
A common method for finding $\pi_\text{opt}$ is the PPO2 algorithm\cite{schulman2017ppo}, which is an improvement over the Actor-Critic (AC) framework\cite{konda1999ac}, capable of optimizing policy in high-dimensional, nonlinear, and continuous state spaces. The AC algorithm combines policy-based and value-based methods, directly modeling and learning the policy while using the value function as a guide to update the policy.

The AC algorithm uses two neural networks: the Actor network and the Critic network. The Actor network selects actions based on the current state, while the Critic network estimates the value function of the current state. Both networks are trained using the policy gradient method, updating the neural network weights using gradient descent.

The PPO2 algorithm also utilizes a policy network and a value network for policy optimization but introduced a clipped surrogate objective function: 
\begin{equation} \label{ppo}
    {\mathcal{L}^{clip}(\theta)} =: \mathop{\mathbb{E}_{t}\left[\min(\mu_t(\theta)\hat{A}_t, \operatorname{clip}_{\epsilon}(\mu_t(\theta), 1-\epsilon, 1+\epsilon)\hat{A}_t))\right]}
\end{equation}
where the $\hat{A}_t$ is the estimated advantage function, $\operatorname{clip}_{\epsilon}$ is the clip operator, $\epsilon$ is the clipping parameter. For each epoch, the PPO2 algorithm updates the process in several steps:

\begin{enumerate}
    \item Sample and collect experience data. In this step, it uses the old policy network to sample data and calculate the ratio $\mu_t(\theta) = \frac{\pi(A_t|S_t)}{\pi_\text{old}(A_t|S_t)}$ between the old and new policy networks at each state to control the step size of the policy update. Meanwhile, it uses the value function network (Critic) to estimate the advantage function at each state. For each collected state-action pair $(s_t, a_t)$, compute the corresponding reward $R_t$ and the Generalized Advantage Estimation (GAE) $\hat{A}_t(\lambda)$.

    \item Update the new policy network and value network. Select a certain batch of trajectory samples from the collected trajectories and compute the policy network and value network outputs under these state samples. Update the policy network parameters using stochastic gradient descent to minimize the objective function $-L(\theta)$. The expression for $L(\theta)$ is as follows:

\begin{equation}
    L(\theta) = \mathbb{E}_t \left[ \min (a_t, b_t) \right] + c * H(\pi),
\end{equation}

where $a_t = \mu_t(\theta) \hat{A}_t(\lambda)$, $b_t = \operatorname{clip}_\epsilon\left(\mu_t(\theta) \right) \hat{A}_t(\lambda)$, $\mathbb{E}_t$ denotes the expectation. $c$ is the entropy coefficient, and $H(\pi)$ is the policy entropy. The clipping function $\operatorname{clip}_\epsilon$ is defined as:

\begin{equation}
    \operatorname{clip}_\epsilon(x) = \operatorname{clamp}(x, 1 - \epsilon, 1 + \epsilon).
\end{equation}

This clip function limits the magnitude of the policy update within an acceptable range, ensuring policy stability. Meanwhile, PPO2 uses policy entropy to measure the exploratory nature of the policy and adds the negative value of the policy entropy as an additional penalty term to the policy objective function to encourage more exploratory policies and avoid local optima.

\end{enumerate}

\section{MULTI-STAGE REINFORCEMENT LEARNING}
In this work, the tasks of UGV and UAV are not independent, as UGV needs to navigate to the target by observing the relative position of the UAV and following the UAV. The precondition for UGV to navigate to the target location is that UAV must be able to search and move to the target location. If both UAV and UGV agents are trained simultaneously, UGV cannot receive effective reward information and observation information related to UAV before UAV achieves satisfactory training results. Therefore, inspired by the concept of curriculum learning, this paper proposes a multi-stage reinforcement learning method by decomposing the task of the agents. The specific process is as follows: first, we train UAV, and after the policy of the UAV can correctly search for the target, train the policy of the UGV. At the same time, the UAV in the environment continues to train the previously trained model to infer and execute actions, and UGV can obtain effective experience and conduct training with the guidance of the UAV. The proposed multi-stage reinforcement learning algorithm is described in Algorithm \ref{alg:multistage}.

\begin{algorithm}
\caption{Multi-stage Reinforcement Learning for the UAV-UGV System}\label{alg:multistage}
\begin{algorithmic}[1]
\State \textbf{Stage 1}
\State Initialize the environment, keep the UGV stationary, and start training the UAV
\State $reward\_counter \gets 0$
\While{$reward\_counter < 50$}
    \State $reward\_sum \gets 0$
    \For{$i \in \{1, \dots, 10000\}$}
        \State Interact with the environment using the UAV
        \State Update the policy and value function of the UAV
        \State $reward\_sum \gets reward\_sum +$ $R_t$
    \EndFor
    \If{$\frac{reward\_sum}{10000} \geq 5000$}
        \State $reward\_counter \gets reward\_counter + 1$
    \Else
        \State $reward\_counter \gets 0$
    \EndIf
\EndWhile
\State \textbf{Stage 2}
\State Initialize the environment, continue training the UAV while starting to train the UGV
\State $step \gets 0$
\While{$step < max\_step$}
    \State Interact with the environment using both the UAV and UGV
    \State Update the policy and value functions of the UAV and UGV
    \State $step \gets step + 1$
\EndWhile
\State Stop training
\end{algorithmic}
\end{algorithm}

\subsection{Training Settings of UAV}
\subsubsection{Action Space}
% In the case of ignoring air resistance and wind disturbance, the dynamic model of a quadrotor can be described as follows, 
In this section, we describe the action space of a quadrotor unmanned aerial vehicle (UAV). Ignoring air resistance and wind disturbances, the dynamics of a quadrotor UAV can be described by the following equations:

\begin{equation}
\frac{d\mathbf{p_w}}{dt} = \mathbf{v_w}
\end{equation}

\begin{equation}
\frac{d\mathbf{v_w}}{dt} = \mathbf{R}\begin{bmatrix} 0 \\ 0 \\ -\frac{T_t}{m} \end{bmatrix} + \begin{bmatrix} 0 \\ 0 \\ g \end{bmatrix}
\end{equation}

\begin{equation}
\frac{d\mathbf{R}}{dt} = \mathbf{R} \mathbf{W_b}^*
\end{equation}

where $\mathbf{p_w}$ and $\mathbf{v_w}$ denote the position vector and linear velocity vector of the UAV in the world coordinate frame ${O_w}$, respectively. The world coordinate frame follows the left-hand rule and is ordered in the $x$, $y$, and $z$ axes, with the $z$ axis pointing in the direction of the gravitational acceleration $g$. The rotation matrix $\mathbf{R}$ represents the attitude of the body coordinate frame ${O_b}$ relative to ${O_w}$. The body angular velocity vector $\mathbf{W_b}=[w_x,w_y,w_z]^T$ represents the roll, pitch, and yaw rates around the $x$, $y$, and $z$ axes, respectively. The matrix $\mathbf{W_b}^*$ is defined as the following skew-symmetric matrix:
\begin{equation}
\mathbf{W_b}^* = \begin{bmatrix} 0 & -w_z & w_y \\ w_z & 0 & -w_x \\ -w_y & w_x & 0 \end{bmatrix}
\end{equation}
By taking the yaw, pitch, and roll angular velocities $[w_x,w_y,w_z]$, as well as the thrust $T_t$ as control variables, and considering the continuous motion control of the UAV in this paper, the output of the action space is designed to be four-dimensional and normalized to the range $[-1,1]$. To save resources, the attitude variation of the UAV during flight, such as the inclination of the UAV when moving forward and backward, is ignored in the simulation environment.

\subsubsection{Observation Space}
The agent selects actions based on the information from the observation space and the trained policy. The decision-making process of the UAV primarily relies on LiDAR raw data. The LiDAR system on the UAV features two different types of laser beams, with each beam having distinct directions and purposes. One laser beam is parallel to the UAV's own attitude direction and is primarily used for obstacle avoidance. Another laser beam has an angle with the UAV's attitude direction and is mainly utilized for scanning the ground and searching for targets. Assuming that the LiDAR mounted on the UAV has target detection capabilities, it can identify the appearance of search and rescue targets. The UAV also has access to its own spatial coordinate information, self-orientation information represented by quaternions, and the continuous action values from the previous step to have a smooth trajectory.

\subsubsection{Reward Function Design}
The UAV receives the following rewards in different situations:
\begin{equation}
R_{\text{uav}}(s_{\text{uav}},a_{\text{uav}}) = 
\begin{cases}
    r_{\text{arrive}} & \text{if } d_t < \theta_1 \\
    r_{\text{collision}} & \text{if collision occurs} \\
    r_{\text{forward}} & \text{if } x_{t-1} < x_t \text{ and } x_t < x_{\text{cross}} \\
    \alpha \cdot \frac{d_{\text{cross}} - d_t}{d_{\text{cross}}} & \text{if } x_t > x_{\text{cross}} \\
    r_{\text{time}} & \text{for each time step}
\end{cases}
\end{equation}

There are four situations that lead to the UAV receiving rewards or penalties, all of which are used in the PPO2 algorithm without normalization or clipping. If the UAV's distance $d_t$ to the target on the plane is less than the threshold $\theta_d$, a positive reward $r_{\text{arrive}}$ is given. However, if the UAV collides with an obstacle, a negative reward $r_{\text{collision}}$ is applied, and the episode ends when either of these conditions is met. A smaller positive reward $r_{\text{forward}}$ is given whenever the UAV moves forward, with the condition that $x_{t-1} < x_t$ and $x_t < x_{\text{cross}}$. Additionally, to inspire the UAV to approach the target, if $x_t > x_{\text{cross}}$, a positive reward proportional to the ratio of $(d_{\text{cross}} - d_t)$ to the maximum distance $d_{\text{cross}}$ at the center of the intersection is arranged, multiplied by a hyperparameter $\alpha$. Furthermore, a negative reward $r_{\text{time}}$ is given at each time step to prompt the UAV to reach the target more quickly.

\subsection{Training Settings of UGV}
\subsubsection{Action Space}
In order to prevent the autonomous vehicle from getting stuck in local optima during the training process, this paper discretizes the vehicle's actions. In the simulation environment, the autonomous vehicle needs to continuously move deeper into the mine to complete the search and rescue mission. In this case, only forward movement is an effective action; thus, discretizing the action space can contribute to more efficient training. The autonomous vehicle has eight available discrete actions, composed of three levels of linear speed and three direction choices (i.e., turning left, no turning, and turning right). The discretized action space is shown in Table 1. In real life, the control of autonomous vehicles should be continuous. Although discretizing the actions of the vehicle may affect its applicability in the real world, in specific environments, this discretization method can improve the performance and efficiency of the autonomous vehicle.
\begin{table}[ht]
\caption{Discretized Action Space for UGV (-1 for left / 1 for right)}
\centering
\begin{tabular}{c c c}
\toprule
Discrete Action ID & Speed Magnitude & Turning Direction  \\
\midrule
1 & 1.5 & -1 \\
2 & 1.5 & 0 \\
3 & 1.5 & 1 \\
4 & 3 & -1 \\
5 & 3 & 0 \\
6 & 3 & 1 \\
7 & 0.75 & -1 \\
8 & 0.75 & 1 \\
\bottomrule
\end{tabular}

\label{table:discretized_action_space}
\end{table}

\subsubsection{Observation Space}
In the decision-making process of the UGV, it not only needs to perform obstacle avoidance based on the information from the LiDAR sensor but also needs to execute the following behavior according to the relative position and attitude information with respect to the UAV. The LiDAR sensor of the UGV only returns the distance to obstacles and does not include target detection functionality. The LiDAR beams extract data every 10 degrees from -90 to 90 degrees, resulting in a total of 19-dimensional LiDAR data. The UGV can obtain the relative position vector and corresponding angular measurements with the UAV on the two-dimensional plane, which can be used by the UGV to determine the following strategy in the next step. Furthermore, the UGV is aware of its ego position, orientation, and the value of the discrete action taken in the previous step.

\subsubsection{Reward Design}
The designed reward functions for UGV is described as follows:
\begin{equation} \label{UGV_reward}
  R_{\text{ugv}}(s_{\text{ugv}}, a_{\text{ugv}}) =
  \begin{cases}
    r_{\text{arrive}} & \text{if } d_t < \theta_1 \\
    r_{\text{collision}} & \text{if collision occurs} \\
    r_{\text{distance}} & \text{if } d_{\text{to\_uav}} < \theta_2 \\
    r_{\text{follow}} & \text{if } x_{\text{ugv}} > x_{\text{uav}} \\
    c_{r1} \frac{x_{\text{ugv}}}{x_{\text{cross}}} & \text{if } x_{\text{ugv}} < x_{\text{cross}} \\
    c_{r1} + c_{r2} \frac{d_{\text{cross}} - d_{\text{t}}}{d_{\text{cross}}} & \text{if } x_{\text{ugv}} > x_{\text{cross}} \\
    r_{\text{time}} & \text{for each time step}
  \end{cases}
\end{equation}

The reward function for the UGV is relatively complex, incorporating various essential rewards to enhance the agent's performance. As shown in (\ref{UGV_reward}), where $d$ represents the distance to the target, $x$ represents the x-axis coordinate value, and $\theta$ and $c$ are distinct constants. In addition to the conventional rewards of $r_{\text{arrive}}$, $r_{\text{collision}}$, and the time penalty $r_{\text{time}}$, the UGV is awarded rewards related to maintaining formation with the UAV. When the distance between the UGV and UAV on the two-dimensional plane is less than $\theta_2$, the UGV receives a positive reward, $r_{\text{distance}}$. Moreover, when the x-coordinate of the UGV is greater than that of the UAV, a negative reward $r_{\text{follow}}$, is assigned to encourage the UGV to learn to follow the UAV and maintain formation. Furthermore, to guide the UGV to move forward and approach the target, a progress reward is calculated based on the distance traveled. If the UGV has passed the intersection, an additional reward is given based on the distance to the target. The calculation method is similar to the rewards for the UAV, which encourages approaching the target, both being determined by the ratio of the total length to the distance the UGV has traveled.

\subsection{Intrinsic Curiosity Module (ICM)}
In the designed mine environment, UAVs need to explore deep scenes, which can lead to sparse reward problems\cite{zhu2021zongshu}. The learning process for the agent may become extremely challenging under such circumstances, and most RL algorithms struggle to achieve satisfactory training results in complex tasks. In a long, obstacle-rich environment, search and navigation tasks can be difficult if relying solely on random exploration. An effective solution is to introduce an intrinsic reward that does not depend on environmental information, similar to biological curiosity\cite{burda2018icm}. The agent receives intrinsic rewards when exploring new environments or discovering new things, hence referred to as the curiosity module. The intrinsic reward signals generated by the agent's curiosity module enable them to explore the environment more proactively, such as discovering different routes to reach the target. The agent's objective function for policy updates combines intrinsic rewards and external rewards through weighted summation.

A standard Intrinsic Curiosity Module (ICM) consists of three neural networks: the feature extractor network, the forward network, and the inverse network. The structure is shown in Fig. 2.
\begin{figure}[htbp]
  \centering
  \includegraphics[width=3.4in]{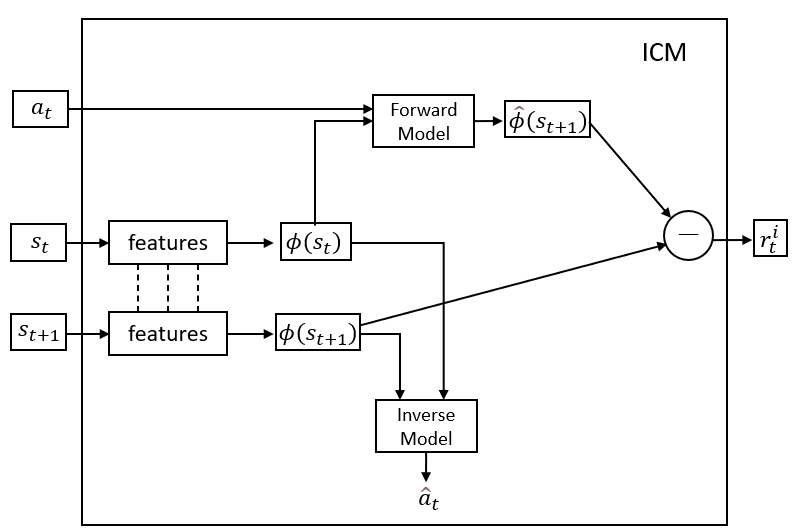}
  \caption{Architecture for the ICM.}
  \label{fig2}
\end{figure}

The feature extractor can be seen as an encoder, which transforms the original state \(s\) into a low-dimensional feature vector \(\varphi(s)\), creating a feature space that only contains information that the agent can control and change. This makes the model robust to environmental changes that the agent cannot control during training. The parameter update of the feature extractor is achieved by minimizing the loss functions of the forward network and inverse network simultaneously.

The forward network aims to predict the next state \(s_{t+1}\) given the current state \(s_t\) and the action \(a_t\). The prediction error between the predicted value and the true value serves as the curiosity reward. The expression for the forward model is:
\begin{equation}
    \hat{\varphi}(s_{t+1}) = f(\varphi(s_t), a_t; \theta_F)
\end{equation}
where \(\varphi(s_t)\) represents the feature representation of the current state \(s_t\), \(a_t\) is the action taken by the agent in state \(s_t\), and \(\theta_F\) is the network parameters of the forward model. The loss function of the forward model is:
\begin{equation}
    L_F = \frac{1}{2} \lVert \hat{\varphi}(s_{t+1}) - \varphi(s_{t+1}) \rVert^2
\end{equation}

The inverse network aims to predict the action the agent needs to take to transit the state \(s_t\) to \(s_{t+1}\). The expression for the inverse model is:
\begin{equation}
    \hat{a}_t = g(\varphi(s_t), \varphi(s_{t+1}); \theta_I)
\end{equation}

where \(\varphi(s_t)\) and \(\varphi(s_{t+1})\) represent the feature representations of the current state \(s_t\) and the next state \(s_{t+1}\), respectively, and \(\theta_I\) is the network parameters of the inverse model. The loss function of the inverse model is:
\begin{equation}
    L_I = -\sum[P(a_{t}) \cdot \log(q(a_{t}))]
\end{equation}

\begin{figure}[htbp]
  \centering
  \includegraphics[width=3.4in]{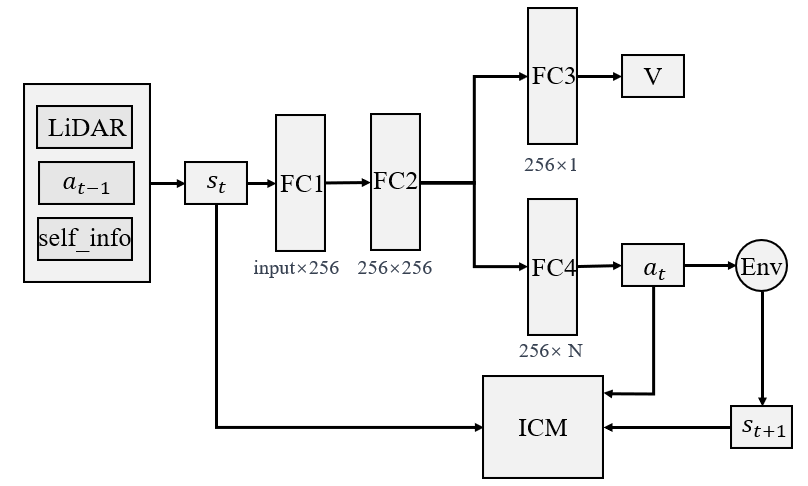}
  \caption{Architecture of the PPO2-ICM model for the policy training.}
  \label{fig3}
\end{figure}

Based on these formulas, the loss functions of the forward model and the inverse model can be calculated, and the network parameters of the feature extractor, forward model, and inverse model can be updated. We employ the PPO2 algorithm combined with the ICM module as the basic algorithm for training the heterogeneous robotic systems. The architecture of the PPO2-ICM model is illustrated in Fig. \ref{fig3}.

We concatenate all observations into a sequence and feed it directly into a fully connected neural network with two hidden layers, each containing 256 neurons. The policy network and the action network share hidden layers but use different output layers. The output dimension $N$ of the policy network depends on the design of the agent's action space.

In our fully connected neural network, a gating mechanism is employed to regulate the information flow of input data. Firstly, we pass the input data through a Sigmoid activation function. Then, we perform element-wise multiplication between the output of the Sigmoid function (ranging from 0 to 1) and the original input data to achieve the gating function. The closer the output value of the Sigmoid function is to 1, the more intact the corresponding original data element is preserved; the closer the output value is to 0, the more severely the corresponding original data element is suppressed. This gating mechanism allows the network to dynamically adjust the weights of individual data elements based on the output of the Sigmoid function, enabling the network to learn selective and adaptive information routing and thereby better capture complex features and relationships from the data.

For the ICM module, the structure of the forward model and the inverse model is similar to the aforementioned fully connected neural network structure, and the output dimension of the encoder is set to 128.

\section{EXPERIMENTS AND RESULTS}
In this section, we introduce the training process, inference testing, and comparative studies of the model. The training process of the model is implemented in a virtual mine environment simulated by Unity3D, and the communication between the DRL algorithm and the simulation environment is facilitated using the ML-Agents toolkit\cite{juliani2018unity}. As shown in Fig \ref{fig1}, the simulation environment features a simple quadrotor UAV model and a standard wheeled UGV. In each episode, the target randomly appears on one of the two intersecting roads.

\subsection{Parameter Settings}
The hyperparameters of the PPO2-ICM algorithm and rewards are listed in Table \ref{table:hyperparameters}. Note that $r_\text{time} = -0.1$. The training tool versions are as follows: mlagents-toolkits: 0.28; ml-agents-envs: 0.28.0; communicator API: 1.5.0; PyTorch: 1.7.1+cu110. The training process was carried out on an Intel Core i7-10875H CPU. In the development of an environment combining Unity and ml-agents-toolkit, the training can be accelerated by replicating multiple environment instances within the same scene. During the training process of this model, there are 30 environment instances within a single scene, which allows for faster sample collection and policy updates. In contrast, when using DRL algorithms for training in Gazebo, typically only a single environment can be run, while the simulation platform presented in this paper, which utilizes Unity3D and ml-agent toolkit, can significantly improve training speed. This is one of the advantages of using a simulation platform based on Unity3D and ml-agent toolkit.
\begin{table}[ht]
\centering
\caption{Hyperparameters used in Multi-stage RL}
\begin{tabular}{l l l l}
\hline
Parameter & UAV & UGV & Reward\\
\hline
learning\_rate & 0.0003 & 0.0002 &  $r_\text{arrive}(\text{UAV})=7000$\\
beta & 0.03 & 0.03 &  $r_\text{collision}(\text{UAV})=-7000$\\
epsilon & 0.2 & 0.3 &  $r_\text{forward}(\text{UAV})=0.5$ \\
lambd & 0.95 & 0.95 &  $\alpha(\text{UAV})=5000$\\
learning\_rate\_schedule & linear & linear 
 & $r_\text{collision}(\text{UGV})=-15000$\\
extrinsic\_gamma & 0.99 & 0.99 & $r_\text{distance}(\text{UGV}) = 5$ \\
extrinsic\_strength & 1.0 & 1.0 & $r_\text{follow}(\text{UGV}) = -20$ \\
curiosity\_gamma & 0.99 & 0.99 & $c_{r1}(\text{UGV}) = 4000$\\
curiosity\_strength & 0.02 & 0.05 & $c_{r1}(\text{UGV}) = 8000$ \\
curiosity\_learning\_rate & 0.0003 & 0.0003 & $r_\text{arrive}(\text{UGV})=3000$ \\
\hline
\end{tabular}
\label{table:hyperparameters}
\end{table}

\subsection{Baseline and Ablation}

\begin{figure}[htbp]
  \centering
  \includegraphics[width=3.4in]{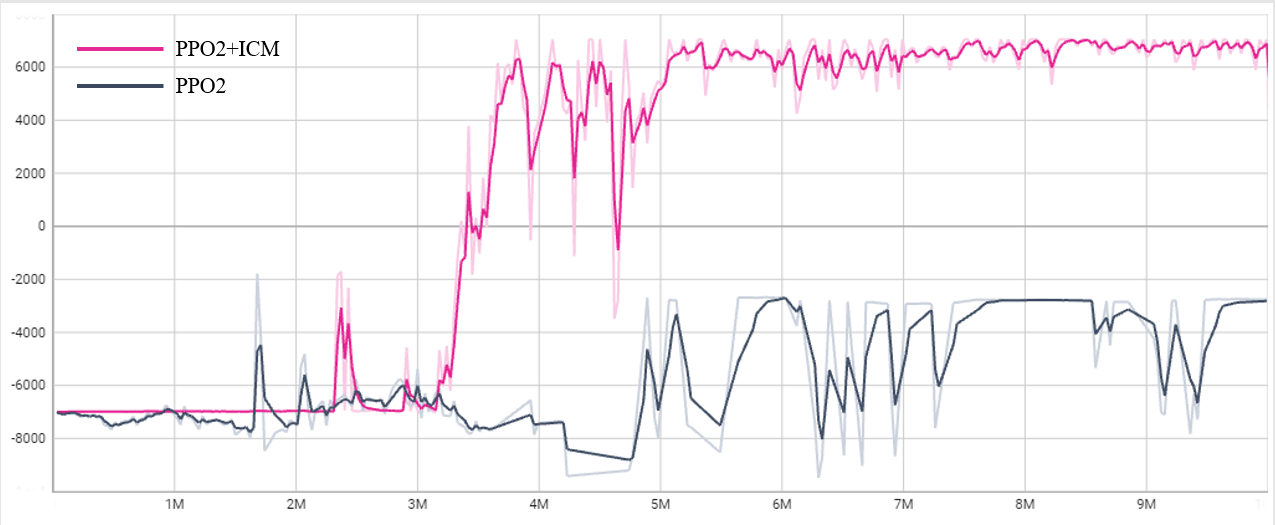}
  \caption{Cumulative reward for the model with/without ICM module. Pink is with the ICM module while black is without the ICM module.}
  \label{fig4}
\end{figure}

In order to evaluate the superiority of the PPO2-ICM algorithm, we compared the performance of PPO2 (without the ICM module), the SAC algorithm and our method. All comparative tests were conducted only during the first stage of training (for UAV training). For the SAC algorithm, its training efficiency was significantly lower compared to the PPO2 algorithm since the target is not stationary. The SAC training was five times slower than PPO2 for the same number of steps, and after testing for one million steps, the reward obtained by the UAV did not show a significant improvement. Fig. \ref{fig4} shows the training results with and without the ICM module. Without the ICM module, UAV was unable to achieve effective training after ten million steps. However, the PPO2 algorithm with the ICM module was able to successfully train the UAV.

\subsection{Multi-stage Reinforcement Learning Experiment}
\begin{figure}
    \centering
    \subfigure[]{
        \includegraphics[width=3.4in, keepaspectratio]{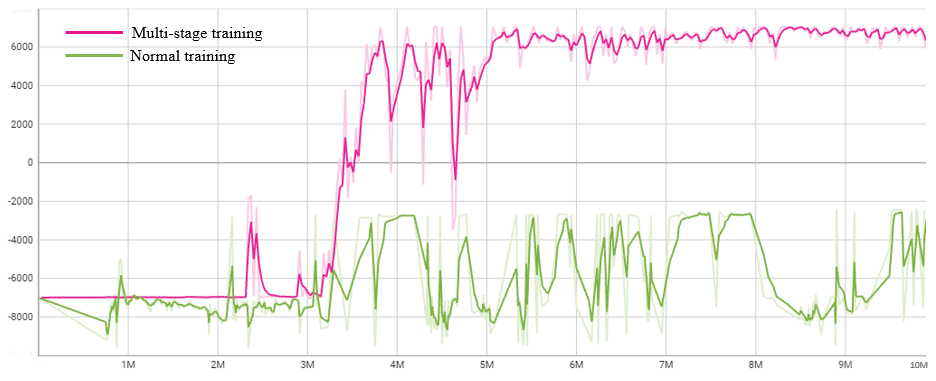}
        \label{fig:4a}
    }
    \\
    \subfigure[]{
        \includegraphics[width=3.4in, keepaspectratio]{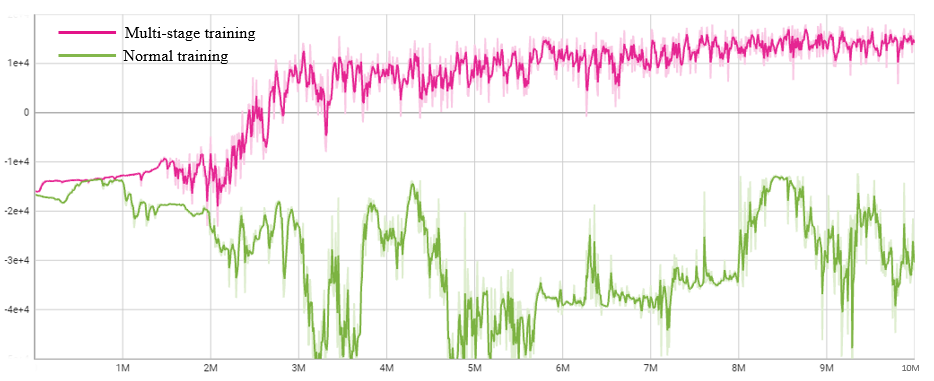}
        \label{fig:4b}
    }
    \caption{ (a) Cumulative reward of UAV. (b) Cumulative reward of UGV.}
    \label{fig:5}
\end{figure}

To assess whether the multi-stage reinforcement learning framework can effectively improve the training performance of the aerial-ground robot system for collaborative navigation problems, we conducted experiments and recorded the results when training both the UAV and UGV models simultaneously in the environment, as shown in Fig. \ref{fig:5}. After \(1 \times 10^7\) steps, it can be observed from the figure that when training both agents concurrently, neither of them can achieve satisfactory training outcomes. This is due to the fact that when training both agents simultaneously, they are unable to receive sufficiently effective reward signals, and there are no effective trajectories to promote the evolution of their policies. However, by employing the multi-stage learning approach, the UGV can obtain valid observations from the relative position information with the UAV during the training process, which is useful for training their collaborative behavior.

\begin{figure}[htbp]
  \centering
  \includegraphics[width=3.4in]{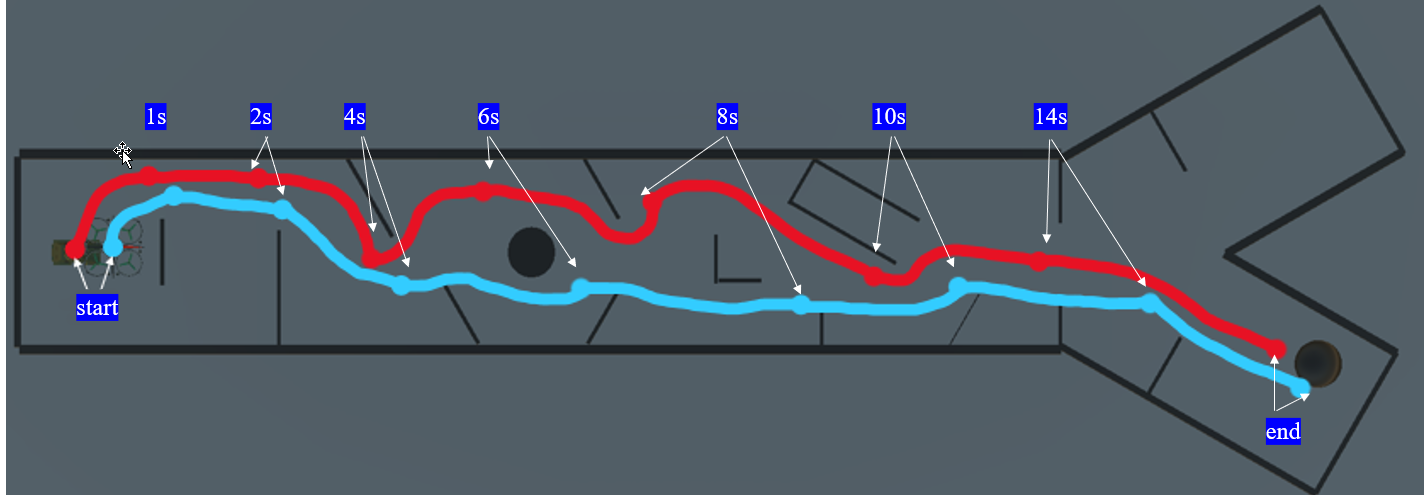}
  \caption{Motion trajectories of the UAV and UGV.}
  \label{fig6}
\end{figure}

\subsection{Inference Experiment}
To evaluate the effectiveness of the proposed model for executing collaborative navigation tasks in a heterogeneous robot system, inference tests were conducted in various simulated environments. In inference mode, the behavior of the UAV and UGV is determined by their policy networks accordingly. In the original training environment, the motion trajectories of the UAV and UGV are shown in Fig. \ref{fig6}. After 1000 episodes, the probability of the UGV successfully reaching the target location is approximately $89.1\%$. During the inference process, the UGV is able to maintain a certain distance behind the UAV and exhibits accelerating chasing behavior when the distance is relatively large. Upon reaching an intersection, the UGV can continuously follow the UAV and move towards the fork where the distressed individuals (targets) are located, even though the UGV's observations do not contain any information related to the target position.

\begin{table}[htbp]
\centering
\caption{Success rate of air-ground robot system}
\label{table:testresults}
\begin{tabular}{c|cc}
\toprule
Algorithms & Original Environment & Complex Environment \\
\midrule
PPO2-ICM &89.1\% & 67.6\%  \\
PPO2 & 0\% & 0\% \\
SAC & 0\% & 0\% \\
\bottomrule
\end{tabular}
\end{table}

Furthermore, we modified the width of the straight path to $\textbf{15~m}$, the angle of the fork to $\textbf{40~degrees}$, the positions of obstacles, and the target angle to create a more complex environment and conducted inference tests within it to test the generalization capability of trained models. The test results are listed in Table \ref{table:testresults}. After 1000 episodes, the success rate was found to be $67.6\%$. Other baseline methods cannot finish the designed task either in the original environment or the complex environment. This demonstrates that despite the environmental randomness during the training process, our policy still shows a certain level of robustness towards the environment. More details can be found in the supplemental video.

\section{Conclusion}
In this paper, we propose a heterogeneous robot collaboration search and navigation strategy based on DRL. We have developed a multi-stage reinforcement learning training framework that sequentially initiates the training of individual agents based on specific tasks. Furthermore, we employ the PPO2 algorithm combined with the ICM module as our training algorithm, providing intrinsic curiosity rewards for the agents to address the sparse reward problem in the environment. From our experiments, compared to conventional RL methods, our approach ensures effective learning of collaborative behaviors among agents, achieving the desired performance and improving the training efficiency of DRL algorithms. Tests in the simulated environment show that the UAV can select the correct route based on laser information, and the UGV, through collaboration with the UAV and solely relying on sparse laser information, can accomplish navigation tasks in completely unknown environments. Our approach and results have practical implications for real-world robot collaborative search and rescue missions.

There is still some further research on the work presented in this paper. We will attempt to incorporate environmental randomness to ensure that the trained policies can be applied to various environments with different shapes and complexities.

% or



\bibliographystyle{ieeetr}
\bibliography{ref}

\begin{thebibliography}{10}

\bibitem{liu2013robotic}
Y.~Liu and G.~Nejat, ``Robotic urban search and rescue: A survey from the
  control perspective,'' {\em Journal of Intelligent \& Robotic Systems},
  vol.~72, pp.~147--165, 2013.

\bibitem{niroui2019deep}
F.~Niroui, K.~Zhang, Z.~Kashino, and G.~Nejat, ``Deep reinforcement learning
  robot for search and rescue applications: Exploration in unknown cluttered
  environments,'' {\em IEEE Robotics and Automation Letters}, vol.~4, no.~2,
  pp.~610--617, 2019.

\bibitem{liu2022review}
C.~Liu, J.~Zhao, and N.~Sun, ``A review of collaborative air-ground robots
  research,'' {\em Journal of Intelligent \& Robotic Systems}, vol.~106, no.~3,
  p.~60, 2022.

\bibitem{gul2019comprehensive}
F.~Gul, W.~Rahiman, and S.~S. Nazli~Alhady, ``A comprehensive study for robot
  navigation techniques,'' {\em Cogent Engineering}, vol.~6, no.~1, p.~1632046,
  2019.

\bibitem{kostrikov2020drlimage}
I.~Kostrikov, D.~Yarats, and R.~Fergus, ``Image augmentation is all you need:
  Regularizing deep reinforcement learning from pixels,'' {\em arXiv preprint
  arXiv:2004.13649}, 2020.

\bibitem{liu2021lifelong}
B.~Liu, X.~Xiao, and P.~Stone, ``A lifelong learning approach to mobile robot
  navigation,'' {\em IEEE Robotics and Automation Letters}, vol.~6, no.~2,
  pp.~1090--1096, 2021.

\bibitem{wu2020towards}
Q.~Wu, X.~Gong, K.~Xu, D.~Manocha, J.~Dong, and J.~Wang, ``Towards
  target-driven visual navigation in indoor scenes via generative imitation
  learning,'' {\em IEEE Robotics and Automation Letters}, vol.~6, no.~1,
  pp.~175--182, 2020.

\bibitem{xiao2022collaborative}
J.~Xiao, P.~Pisutsin, and M.~Feroskhan, ``Collaborative target search with a
  visual drone swarm: An adaptive curriculum embedded multi-stage reinforcement
  learning approach,'' {\em arXiv preprint arXiv:2204.12181}, 2022.

\bibitem{nguyen2020deep}
T.~T. Nguyen, N.~D. Nguyen, and S.~Nahavandi, ``Deep reinforcement learning for
  multiagent systems: A review of challenges, solutions, and applications,''
  {\em IEEE transactions on cybernetics}, vol.~50, no.~9, pp.~3826--3839, 2020.

\bibitem{schulman2017proximal}
J.~Schulman, F.~Wolski, P.~Dhariwal, A.~Radford, and O.~Klimov, ``Proximal
  policy optimization algorithms,'' {\em arXiv preprint arXiv:1707.06347},
  2017.

\bibitem{ding2019messenger}
Y.~Ding, B.~Xin, and J.~Chen, ``Precedence-constrained path planning of
  messenger uav for air-ground coordination,'' {\em Control Theory and
  Technology}, vol.~17, no.~1, pp.~13--23, 2019.

\bibitem{zhang2018uavsensor}
S.~Zhang, H.~Wang, S.~He, C.~Zhang, and J.~Liu, ``An autonomous air-ground
  cooperative field surveillance system with quadrotor uav and unmanned atv
  robots,'' in {\em 2018 IEEE 8th Annual International Conference on CYBER
  Technology in Automation, Control, and Intelligent Systems (CYBER)},
  pp.~1527--1532, IEEE, 2018.

\bibitem{shen2017collaborative}
C.~Shen, Y.~Zhang, Z.~Li, F.~Gao, and S.~Shen, ``Collaborative air-ground
  target searching in complex environments,'' in {\em 2017 IEEE International
  Symposium on Safety, Security and Rescue Robotics (SSRR)}, pp.~230--237,
  IEEE, 2017.

\bibitem{mueggler2014aerial}
E.~Mueggler, M.~Faessler, F.~Fontana, and D.~Scaramuzza, ``Aerial-guided
  navigation of a ground robot among movable obstacles,'' in {\em 2014 IEEE
  International Symposium on Safety, Security, and Rescue Robotics (2014)},
  pp.~1--8, IEEE, 2014.

\bibitem{zhu2017target}
Y.~Zhu, R.~Mottaghi, E.~Kolve, J.~J. Lim, A.~Gupta, L.~Fei-Fei, and A.~Farhadi,
  ``Target-driven visual navigation in indoor scenes using deep reinforcement
  learning,'' in {\em 2017 IEEE international conference on robotics and
  automation (ICRA)}, pp.~3357--3364, IEEE, 2017.

\bibitem{tai2017virtualtoreal}
L.~Tai, G.~Paolo, and M.~Liu, ``Virtual-to-real deep reinforcement learning:
  Continuous control of mobile robots for mapless navigation,'' in {\em 2017
  IEEE/RSJ International Conference on Intelligent Robots and Systems (IROS)},
  pp.~31--36, IEEE, 2017.

\bibitem{marchesini2020discrete}
E.~Marchesini and A.~Farinelli, ``Discrete deep reinforcement learning for
  mapless navigation,'' in {\em 2020 IEEE International Conference on Robotics
  and Automation (ICRA)}, pp.~10688--10694, IEEE, 2020.

\bibitem{schulman2017ppo}
J.~Schulman, F.~Wolski, P.~Dhariwal, A.~Radford, and O.~Klimov, ``Proximal
  policy optimization algorithms,'' {\em arXiv preprint arXiv:1707.06347},
  2017.

\bibitem{konda1999ac}
V.~Konda and J.~Tsitsiklis, ``Actor-critic algorithms,'' {\em Advances in
  neural information processing systems}, vol.~12, 1999.

\bibitem{zhu2021zongshu}
K.~Zhu and T.~Zhang, ``Deep reinforcement learning based mobile robot
  navigation: A review,'' {\em Tsinghua Science and Technology}, vol.~26,
  no.~5, pp.~674--691, 2021.

\bibitem{burda2018icm}
Y.~Burda, H.~Edwards, D.~Pathak, A.~Storkey, T.~Darrell, and A.~A. Efros,
  ``Large-scale study of curiosity-driven learning,'' {\em arXiv preprint
  arXiv:1808.04355}, 2018.

\bibitem{juliani2018unity}
A.~Juliani, V.-P. Berges, E.~Teng, A.~Cohen, J.~Harper, C.~Elion, C.~Goy,
  Y.~Gao, H.~Henry, M.~Mattar, {\em et~al.}, ``Unity: A general platform for
  intelligent agents,'' {\em arXiv preprint arXiv:1809.02627}, 2018.

\end{thebibliography}
\end{document}